\documentclass[pdflatex,sn-vancouver-num]{sn-jnl}
\usepackage{graphicx}
\usepackage{amsmath,amssymb}
\usepackage{booktabs}
\usepackage{array}
\usepackage{placeins}
\usepackage{needspace}
\usepackage{xcolor}

\renewcommand{\abstractfont}{\normalfont\normalsize\leftskip=24pt\rightskip=24pt\parfillskip=0pt plus 1fil}
\renewcommand{\abstractheadfont}{\normalfont\normalsize\bfseries\titraggedcenter}
\renewcommand{\abstractsubheadfont}{\normalfont\normalsize\bfseries}

\unnumbered
\AtBeginDocument{\raggedbottom}
\AddToHook{cmd/maketitle/after}{\setlength{\parskip}{12pt plus 2pt minus 2pt}}

\begin{document}
\title[End-to-end prostate lesion segmentation and detection]{End-to-end prostate lesion segmentation and detection from undersampled diffusion-weighted MRI}

\author*[1,2,3,4]{\fnm{Lukas T.} \sur{Rotkopf}}\email{l.rotkopf@dkfz-heidelberg.de}
\author[2]{\fnm{Marco} \sur{Schlimbach}}
\author[2]{\fnm{Kevin} \sur{Kr\"{o}ninger}}
\author[4]{\fnm{Stefan O.} \sur{Schoenberg}}
\author[3]{\fnm{Heinz-Peter} \sur{Schlemmer}}
\author[1,2,5,6,7,8]{\fnm{Jens} \sur{Kleesiek}}
\author[1,2,5,9]{\fnm{Moritz} \sur{Rempe}}

\affil[1]{\orgdiv{Institute for AI in Medicine (IKIM)}, \orgname{University Hospital Essen}, \orgaddress{\street{Girardetstra\ss{}e 2}, \postcode{45131}, \city{Essen}, \country{Germany}}}
\affil[2]{\orgdiv{Department of Physics}, \orgname{Technical University Dortmund}, \orgaddress{\street{Otto-Hahn-Stra\ss{}e 4a}, \postcode{44227}, \city{Dortmund}, \country{Germany}}}
\affil[3]{\orgdiv{Division of Radiology}, \orgname{German Cancer Research Center (DKFZ)}, \orgaddress{\postcode{69120}, \city{Heidelberg}, \country{Germany}}}
\affil[4]{\orgdiv{Department of Radiology and Nuclear Medicine}, \orgname{University Medical Center Mannheim, Heidelberg University}, \orgaddress{\street{Theodor-Kutzer-Ufer 1-3}, \postcode{68167}, \city{Mannheim}, \country{Germany}}}
\affil[5]{\orgname{Cancer Research Center Cologne Essen (CCCE), University Medicine Essen}, \orgaddress{\street{Hufelandstra\ss{}e 55}, \postcode{45147}, \city{Essen}, \country{Germany}}}
\affil[6]{\orgdiv{Site Essen}, \orgname{RACOON Study Group}, \orgaddress{\city{Essen}, \country{Germany}}}
\affil[7]{\orgdiv{Partner Site Essen}, \orgname{German Cancer Consortium (DKTK)}, \orgaddress{\street{Hufelandstra\ss{}e 55}, \postcode{45147}, \city{Essen}, \country{Germany}}}
\affil[8]{\orgdiv{Medical Faculty and Faculty of Computer Science}, \orgname{University of Duisburg-Essen}, \orgaddress{\postcode{45141}, \city{Essen}, \country{Germany}}}
\affil[9]{\orgdiv{Department of Radiology}, \orgname{Brigham and Women's Hospital, Harvard Medical School}, \orgaddress{\city{Boston}, \country{USA}}}
\abstract{
\paragraph{Background}
Automated prostate lesion detection from diffusion-weighted MR imaging (DWI) conventionally relies on reconstructed images and does not fully use all acquisition information. We present an end-to-end network that uses complex multicoil DWI data and coil sensitivity information for prostate lesion segmentation. We evaluated its robustness to undersampling and added noise, its performance in patient-level detection, and the contributions of data consistency and training strategy.

\paragraph{Methods}
We included 269 fastMRI prostate examinations with additional voxel-level lesion annotations based on the released DWI PIRADS scores. Of these, 64 examinations contained PIRADS 4 or 5 lesions. The proposed network, VarNet-Seg, consists of a k-space variational network (VarNet) with six variational cascades and an image-space 3D segmentation U-Net and is trained end to end on complex multicoil b50 and b1000 data. Comparator models used the same segmentation architecture with fully coil-combined images (CC-Seg), fixed root-sum-of-squares reconstruction (RSS-Seg), or fixed SENSE reconstruction (SENSE-Seg). VarNet-Seg (no DC) retained the proposed architecture with data consistency disabled. We performed ablations with end-to-end training and data consistency enabled or disabled. We evaluated lesion segmentation using Dice and patient-level detection using AUROC in five-fold cross-validation with full sampling, simulated undersampling, added noise, or both.

\paragraph{Results}
VarNet-Seg achieved the highest Dice at every undersampling level. At eight-fold acceleration, Dice was 0.22 for VarNet-Seg, 0.14 for SENSE-Seg and 0.16 for VarNet-Seg (no DC). The paired difference from SENSE-Seg was 0.08 (95\% CI, 0.04--0.11). VarNet-Seg (frozen) and VarNet-Seg (frozen, no DC) achieved Dice of 0.17 and 0.12, respectively, with the matched ablations supporting contributions from both end-to-end training and data consistency. With full sampling, VarNet-Seg and RSS-Seg performed similarly (0.40 vs 0.39). Removing data consistency reduced Dice under acceleration in all training strategies. Patient-level AUROC decreased from 0.75 with full sampling to 0.63 at eight-fold acceleration; corresponding RSS-Seg AUROCs were 0.76 and 0.64. VarNet-Seg inference took 29~ms per volume, compared with 4~ms for SENSE-Seg.

\paragraph{Conclusions}
The proposed multicoil network improved prostate lesion segmentation under simulated undersampling compared to conventional models. Matched ablations supported contributions from both end-to-end training and data consistency. Prospective evaluation is needed to establish whether this approach can be used clinically in accelerated prostate DWI.
}
\maketitle

\section{Introduction}
\label{sec:introduction}

Prostate cancer is one of the most common cancers in men, and its increasing global burden places growing demands on diagnostic services~\cite{James2024}.
Magnetic resonance imaging (MRI) has become central to diagnosis and biopsy planning by identifying suspicious lesions and guiding targeted tissue sampling~\cite{Kasivisvanathan2018,Turkbey2019}. However, long examination times and limited scanner capacity constrain access to prostate MRI~\cite{Ng2025}. Shorter protocols would allow more patients to be examined with existing capacity, provided that they preserve the information needed to detect and localize lesions. For automated lesion assessment, we address the separation of reconstruction and segmentation by training both stages jointly to preserve lesion information under reduced sampling and added noise.

Biparametric protocols already shorten examinations by omitting contrast-enhanced imaging, without loss of diagnostic performance when image quality is adequate~\cite{Ng2025}. They consist of T2-weighted and diffusion-weighted imaging (DWI), including derived apparent diffusion coefficient maps, which provide the anatomical and diffusion information used in PIRADS assessment~\cite{Turkbey2019}.
Further reductions in scan time require more efficient acquisition, for example by collecting fewer k-space samples or reducing the number of signal averages.
Both make the subsequent reconstruction harder, because images must then be recovered from incomplete or noisier measurements.
One common approach, parallel imaging, uses differences in receiver-coil sensitivity to reconstruct undersampled data, but the associated loss of signal-to-noise ratio (SNR) limits acceleration~\cite{Pruessmann1999,Griswold2002}.

A similar constraint arises in low-field MRI, where less expensive hardware could broaden access and modern systems are being investigated for prostate imaging~\cite{Koonjoo2021,Segeroth2025}.
Lower field strength reduces the available MR signal, increasing the relative influence of noise, particularly in high-$b$ DWI where diffusion weighting further attenuates the signal. Additional averaging restores SNR, but at the cost of longer acquisitions~\cite{Koonjoo2021}.

Established approaches to automated prostate lesion assessment from these measurements operate on reconstructed images, with reconstruction and lesion analysis treated as separate tasks~\cite{Saha2024,VanLohuizen2024}.
The analysis network thus receives images after coil combination, without direct access to the underlying complex measurements or the ability to influence their reconstruction.

In contrast, training both reconstruction and segmentation simultaneously allows lesion annotations to guide reconstruction, giving both stages a shared objective~\cite{Sun2019}. Errors in the final segmentation can then influence how the complex diffusion data are combined and refined, so that reconstruction is trained to support lesion delineation. However, improvements in reconstructed image quality do not necessarily translate into better lesion detection~\cite{VanLohuizen2024}. Optimizing reconstruction for the downstream task rather than for image quality addresses this mismatch directly.

Here, we present VarNet-Seg, an end-to-end variational network that combines multicoil DWI reconstruction and 3D prostate lesion segmentation in a single model. Using complex low- and high-$b$ data from the 3~T fastMRI Prostate dataset with additional voxel-level annotations~\cite{Tibrewala2024}, we compare it with segmentation models using fixed RSS or SENSE reconstruction (RSS-Seg and SENSE-Seg), a coil-combined image model (CC-Seg) and VarNet-Seg (no DC), a matched variant with data consistency disabled. Variational networks provide this connection by alternating learned image refinement with data-consistency updates in k-space~\cite{Hammernik2018,Sriram2020}. These updates relate intermediate image estimates to the retained measurements, allowing learned suppression of aliasing and noise to be guided by the input data.

By evaluating full sampling, simulated undersampling, added noise and their combination, we test how far task-trained reconstruction can sustain lesion analysis as data availability and SNR decrease, and whether improved delineation is accompanied by better patient-level detection.
\section{Materials and methods}
\label{sec:methods}
\begin{figure}[htp]
    \centering
    \includegraphics[width=\linewidth]{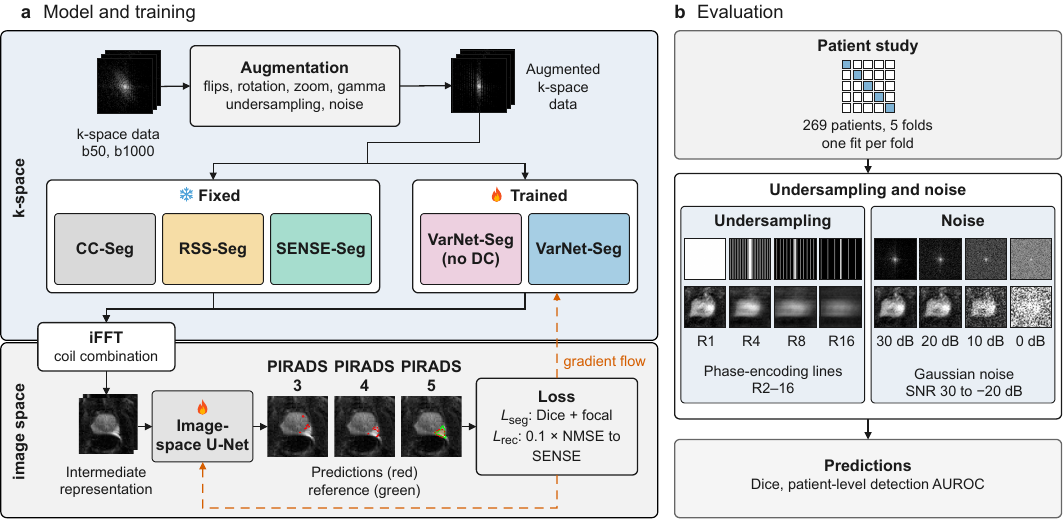}
    \caption{Model architectures and evaluation. (a) The illustrated core architectures use the same 3D segmentation U-Net in image space, with the k-space stage either fixed or trained jointly with segmentation. Red dashed contours indicate class-specific predictions and green contours the reference annotations. (b) Illustration of the evaluation setup with full sampling, undersampling, added noise and their combination. Sampling parameters are provided in Table~\ref{tab:conditions}.}
    \label{fig:method}
\end{figure}
\subsection{Study cohort and reference annotations}

This retrospective study included 269 examinations from the public fastMRI prostate dataset~\cite{Tibrewala2024}. Of the 312 released examinations, 43 were excluded because diffusion raw data were incomplete or missing. Detailed information about the dataset are available in the accompanying publication. In brief, images were acquired on two 3~T systems (MAGNETOM Vida, Siemens Healthineers, Germany) using single-shot echo-planar imaging at $b=50$ and $1000\,\mathrm{s/mm^2}$.
The protocol comprised three diffusion directions, four and twelve averages per direction for the two $b$-values, respectively, two-fold parallel imaging, and $2.0\times2.0\times3.0$~mm voxels. The source study had institutional review board approval with a waiver of informed consent~\cite{Tibrewala2024}.

For this study, an experienced oncological radiologist delineated lesions on the calculated $b1500$ images and apparent diffusion coefficient (ADC) maps, using the dataset's DWI PIRADS ratings as guide to assign lesion grades. PIRADS (Prostate Imaging Reporting and Data System) is a five-point MRI suspicion scale for clinically significant prostate cancer; categories 3, 4 and 5 denote intermediate, high and very high suspicion, respectively~\cite{Turkbey2019}. The ratings used here were assigned to DWI independently of T2-weighted imaging~\cite{Tibrewala2024}. The segmentation classes were background, PIRADS~3, PIRADS~4 and PIRADS~5. Table~\ref{tab:cohort} in the Supplement summarizes the cohort.

\subsection{Data preprocessing}
The diffusion repetitions were regridded for ramp sampling and reconstructed with GRAPPA~\cite{Griswold2002} for viewing. Geometric coil compression (GCC) was used for preprocessing of the k-space data to reduce the physical receive coils to twelve virtual coils for the multicoil models~\cite{Zhang2013}. This number was chosen because they yielded the highest Dice among the evaluated coil counts.
After low-resolution phase correction, complex images were averaged across repetitions and diffusion directions for each $b$-value~\cite{Eichner2015}.
The images were aligned with the reference annotations, cropped to a $100\times100$ matrix and transformed losslessly back to k-space for each coil individually.

ESPIRiT coil sensitivity maps were estimated from the $b50$ data and used for both diffusion contrasts~\cite{Uecker2014}. Each examination was normalized by a single root-mean-square value across coils and $b$-values, preserving their relative amplitudes. For the coil-combined segmentation model (CC-Seg), physical-coil images were combined using the dataset's sensitivity maps, preserving the complex signal, and otherwise underwent the same preprocessing steps. Figure~\ref{fig:method} summarizes the models and evaluation.

\subsection{Network architecture and comparator models}
The proposed model, VarNet-Seg, uses six variational cascades for k-space processing~\cite{Hammernik2018,Sriram2020} with an image-space 3D segmentation U-Net~\cite{Cicek2016} as a single architecture trained end to end for prostate lesion segmentation. The variational module was adapted to process the complex multicoil $b50$ and $b1000$ data jointly. Each cascade uses a 3D residual CNN that receives the real and imaginary components of both contrasts and learns features across neighboring slices. Image refinement alternates with soft data-consistency updates at the sampled k-space locations. Each cascade has separate CNN weights, a learned residual scale and a learned data-consistency weight constrained between zero and one. After SENSE coil combination~\cite{Pruessmann1999} of the processed k-space data, a central crop encompassing the prostate and periprostatic tissue is passed to the image-space U-Net to predict background and the three PIRADS classes.

Six comparator models were chosen to distinguish the contributions of the input representation, reconstruction training strategy and data consistency. CC-Seg receives the complex images reconstructed from the physical receive coils directly. RSS-Seg uses fixed root-sum-of-squares (RSS) magnitude reconstructions from the twelve virtual coils in k-space. SENSE-Seg uses fixed complex SENSE reconstructions from the same virtual coils and sensitivity maps as VarNet-Seg, with missing k-space samples set to zero. VarNet-Seg (no DC) retains the complete reconstruction and segmentation architecture but omits the consistency updates. VarNet-Seg (frozen) and VarNet-Seg (frozen, no DC) use a separately pretrained reconstruction VarNet that is held fixed while the image-space U-Net is trained. Together, the four VarNet variants compare end-to-end and separate training with data consistency enabled or disabled within the same base architecture. All models use the same segmentation U-Net, with input channels adapted to magnitude or complex images. Corresponding segmentation U-Net layers were initialized identically across models within each fold.

\subsection{Training and inference}
Models were trained and evaluated using patient-level five-fold cross-validation stratified by highest annotated lesion grade. All models used the same patient partitions, with a separate validation subset within each training fold. Segmentation training used AdamW with cosine learning-rate decay and early stopping based on validation Dice with full sampling. The same checkpoint was used for all evaluation settings within each fold.

The segmentation loss combined Dice loss~\cite{Milletari2016}, averaged over foreground classes present in each training window, and focal loss~\cite{Lin2017} across all four classes, including background. VarNet-Seg and VarNet-Seg (no DC) also used, as an auxiliary reconstruction loss, the normalized mean squared error between their output magnitudes and the SENSE reference images with full sampling, calculated separately for each $b$-value within the segmentation crop. Models with fixed reconstructions were trained with the segmentation loss alone. Training augmentation included extensive k-space augmentation, namely spatial transformations, gamma adjustment, random or equispaced undersampling at two- to eight-fold acceleration, and added noise. Spatial transformations were applied jointly to coil images, sensitivity maps and labels. Architecture, loss weights and optimization settings are detailed in the Supplement.

For the frozen variants, reconstruction was first pretrained using the per-$b$-value magnitude NMSE alone,  on the training patients of each fold. All reconstruction parameters were then frozen, and a freshly initialized image-space U-Net was trained with the segmentation loss and clean-validation Dice selection described above. For VarNet-Seg (frozen, no DC), data consistency was disabled during both reconstruction pretraining and segmentation training. 

\subsection{Undersampling and noise}

Performance was evaluated with full sampling ($R=1$), simulated undersampling, added noise and their combination. For undersampling, equispaced phase-encoding lines were omitted with a fully sampled central band at accelerations from two- to sixteen-fold, as detailed in Table~\ref{tab:conditions}. This simulates reduced sampling relevant to faster acquisitions; $R$ denotes the nominal retrospective undersampling factor, not a measured scan-time reduction. Sampling patterns were shared across coils, $b$-values and slices within each examination and were identical for all models. Complex Gaussian noise was added to k-space before undersampling, with signal-to-noise ratios (SNRs) from 30 to $-20$~dB. Combined experiments evaluated up to sixteen-fold acceleration at SNRs of 20, 10 and 5~dB. Noise levels were defined using signal power pooled across coils, slices and both $b$-values. Undersampling and noise were applied to the preprocessed k-space, with sensitivity maps and normalization kept as derived from the fully sampled data.

\subsection{Outcomes and statistical analysis}
The primary endpoint was Dice for the combined PIRADS~4 and 5 lesions, averaged over all patients with a lesion in either category.  Grade-specific Dice was calculated among patients with a reference lesion of the corresponding grade. Confusion matrices were normalized by reference class within each patient before averaging across patients with that class.

Patient-level detection was assessed using the area under the receiver operating characteristic curve (AUROC), with predicted volume of the combined lesion region as the patient score. Examinations without a target lesion, including those with PIRADS~3 lesions only, were classified as negative. AUROC was calculated within each test fold and averaged across folds.

Confidence intervals (95\%) were estimated bootstrap resamples stratified by fold and highest reference grade~\cite{Efron1994}. The same resamples were used for all models and evaluation settings, with model differences calculated within each resample. VarNet-Seg--RSS-Seg Dice comparisons used two-sided bootstrap tests with Holm correction within each figure panel~\cite{Holm1979}. Other paired differences were reported with unadjusted confidence intervals.

\FloatBarrier
\Needspace{8\baselineskip}
\section{Results}
\label{sec:results}

\begin{figure}[!htbp]
    \centering
    \includegraphics[width=\linewidth]{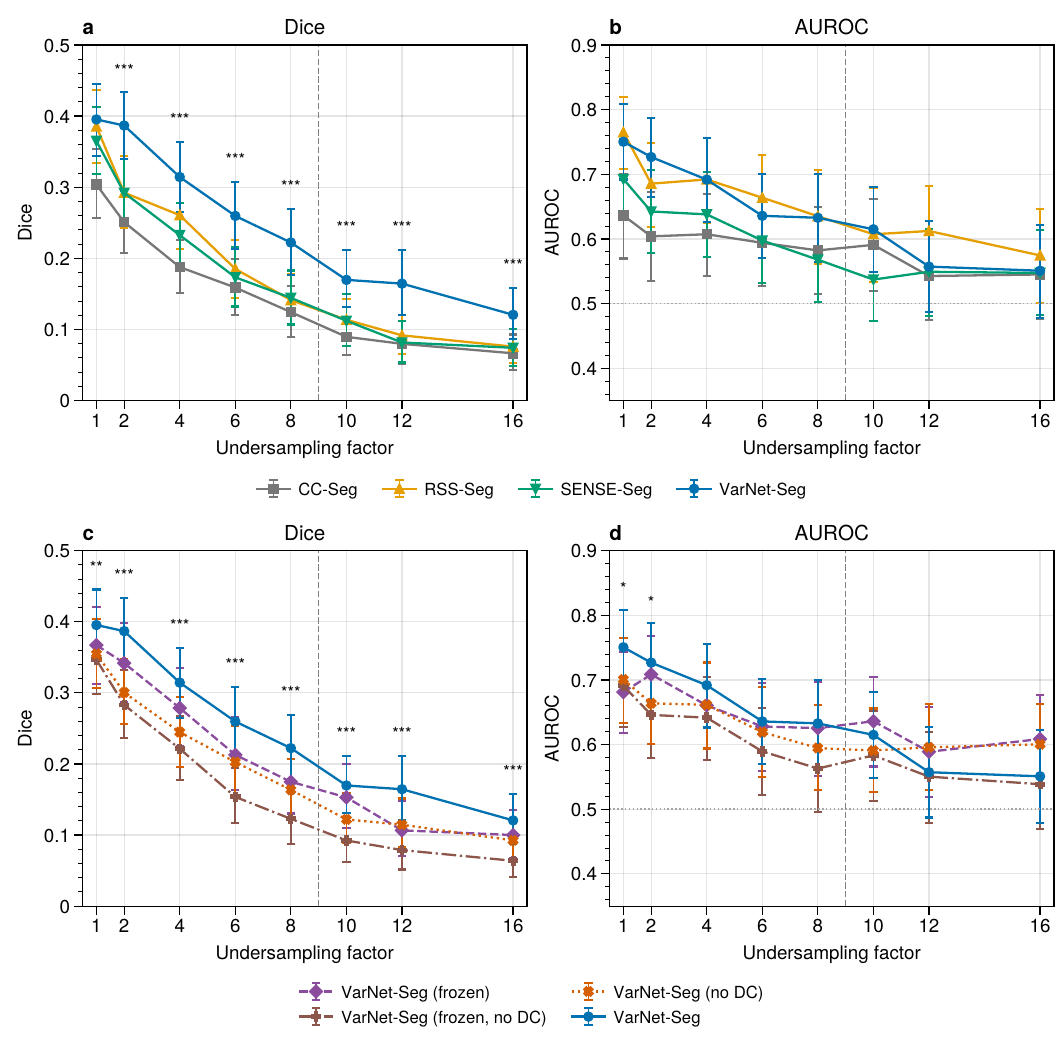}
    \caption{Segmentation and patient-level detection with undersampling. Dice for combined PIRADS~4+5 lesions is shown in the left column (a, c) and patient-level AUROC in the right column (b, d). The upper row compares CC-Seg, RSS-Seg, SENSE-Seg and VarNet-Seg. The lower row compares VarNet-Seg (frozen), VarNet-Seg (frozen, no DC), VarNet-Seg (no DC) and VarNet-Seg. Error bars show 95\% bootstrap confidence intervals. Asterisks indicate VarNet-Seg--RSS-Seg comparisons in a and VarNet-Seg--VarNet-Seg (frozen, no DC) comparisons in c and d ($^{*}p<0.05$, $^{**}p<0.01$, $^{***}p<0.001$). The dashed vertical line separates the training range from higher undersampling factors.}
    \label{fig:masking}
\end{figure}

\subsection{Segmentation and detection performance under undersampling}
VarNet-Seg achieved the highest Dice at every undersampling factor, with a Dice of 0.395 (95\% CI, 0.344--0.445) at full sampling. Dice was 0.385 (95\% CI, 0.334--0.437) for RSS-Seg and 0.304 (95\% CI, 0.257--0.354) for CC-Seg, as shown in Figure~\ref{fig:masking}a. Confidence intervals for the paired differences from SENSE-Seg and VarNet-Seg (no DC) excluded zero, whereas the difference from RSS-Seg was not significant.

At a simulated k-space undersampling factor of eight ($R=8$), Dice was 0.222 for VarNet-Seg and 0.123--0.175 for the other multicoil models, with a paired difference of 0.078 from SENSE-Seg (95\% CI, 0.043--0.113). Dice decreased with increasing acceleration in all models, reaching 0.121 for VarNet-Seg at sixteen-fold acceleration. VarNet-Seg (no DC) remained below VarNet-Seg throughout.

For patient-level lesion detection, VarNet-Seg AUROC decreased from 0.750 with full sampling to 0.633 at eight-fold acceleration and approached chance level at twelve-fold acceleration, as shown in Figure~\ref{fig:masking}b. Corresponding RSS-Seg AUROCs were 0.765 with full sampling and 0.635 at eight-fold acceleration.

\begin{figure}[!htbp]
    \centering
    \includegraphics[width=\linewidth]{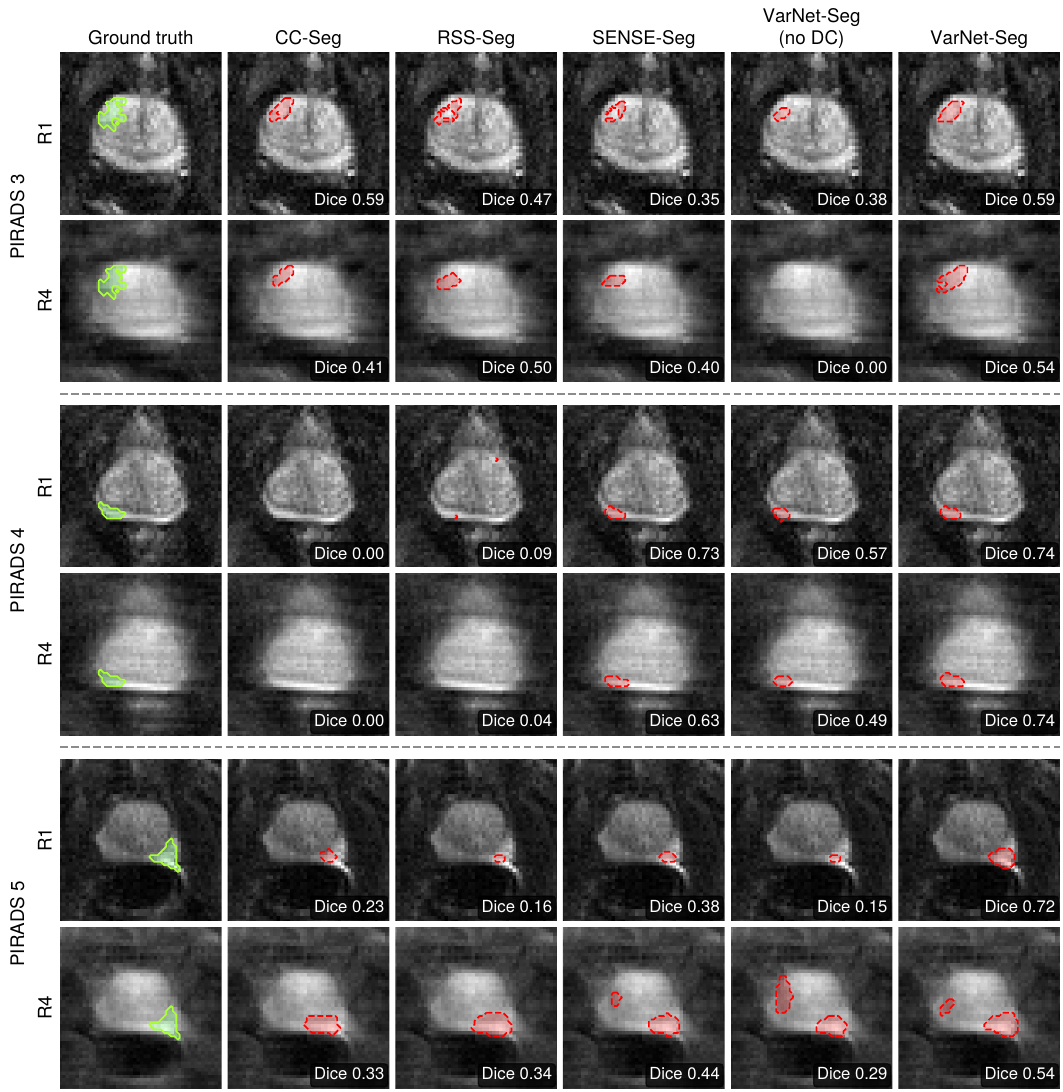}
    \caption{Examples of PIRADS~3, 4 and 5 lesions with full sampling and four-fold undersampling. Contours show the indicated grade only: reference annotations in green and predictions as red dashed lines. Background images are SENSE $b1000$ reconstructions, with matching slices and crops across models. Dice values describe the displayed grade over the full prediction volume.}
    \label{fig:examples}
\end{figure}

Segmentation examples with full sampling and four-fold undersampling are shown in Figure~\ref{fig:examples}.

\subsection{Effects of reconstruction training and data consistency}
The four VarNet variants are compared in Figure~\ref{fig:masking}c,d. At $R=8$, Dice was 0.222 for VarNet-Seg, 0.175 for VarNet-Seg (frozen), 0.164 for VarNet-Seg (no DC) and 0.123 for VarNet-Seg (frozen, no DC). End-to-end training increased Dice relative to the corresponding frozen frontend both with data consistency enabled (95\% CI, 0.009--0.085) and disabled (95\% CI, 0.019--0.065). Enabling data consistency increased Dice under end-to-end training by 0.059 (95\% CI, 0.027--0.091) and under separate reconstruction and segmentation training by 0.052 (95\% CI, 0.027--0.081).

\subsection{Performance with added noise}
\begin{figure}[!htbp]
    \centering
    \includegraphics[width=\linewidth]{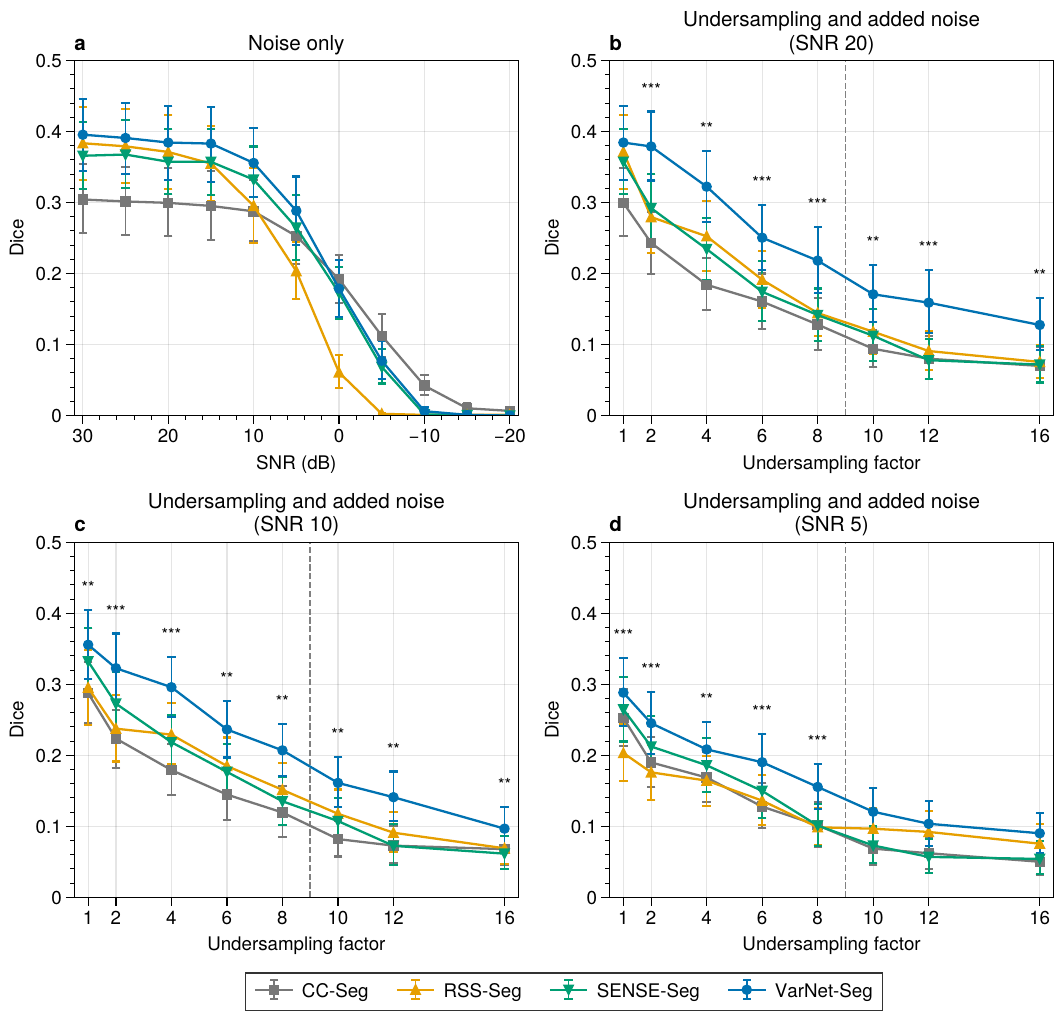}
    \caption{Overall segmentation Dice with added noise for CC-Seg, RSS-Seg, SENSE-Seg and VarNet-Seg. (a) Noise only. (b--d) Undersampling and added noise at SNRs of 20, 10 and 5~dB, respectively. Asterisks in b--d indicate VarNet-Seg--RSS-Seg comparisons ($^{**}p<0.01$, $^{***}p<0.001$). }
    \label{fig:noise}
\end{figure}

In the noise experiments without undersampling shown in Figure~\ref{fig:noise}a, VarNet-Seg achieved the highest Dice at SNRs of 5~dB and above. At 0~dB and below, CC-Seg achieved the highest Dice, while Dice for RSS-Seg approached zero. Figure~\ref{fig:noise}b--d shows that VarNet-Seg also achieved the highest Dice when undersampling was combined with noise. At eight-fold acceleration and an SNR of 5~dB, Dice was 0.156 for VarNet-Seg, compared with 0.101 for SENSE-Seg and 0.099 for RSS-Seg.

\subsection{Lesion grade classification and segmentation errors}

\begin{figure}[!htbp]
    \centering
    \includegraphics[width=\linewidth]{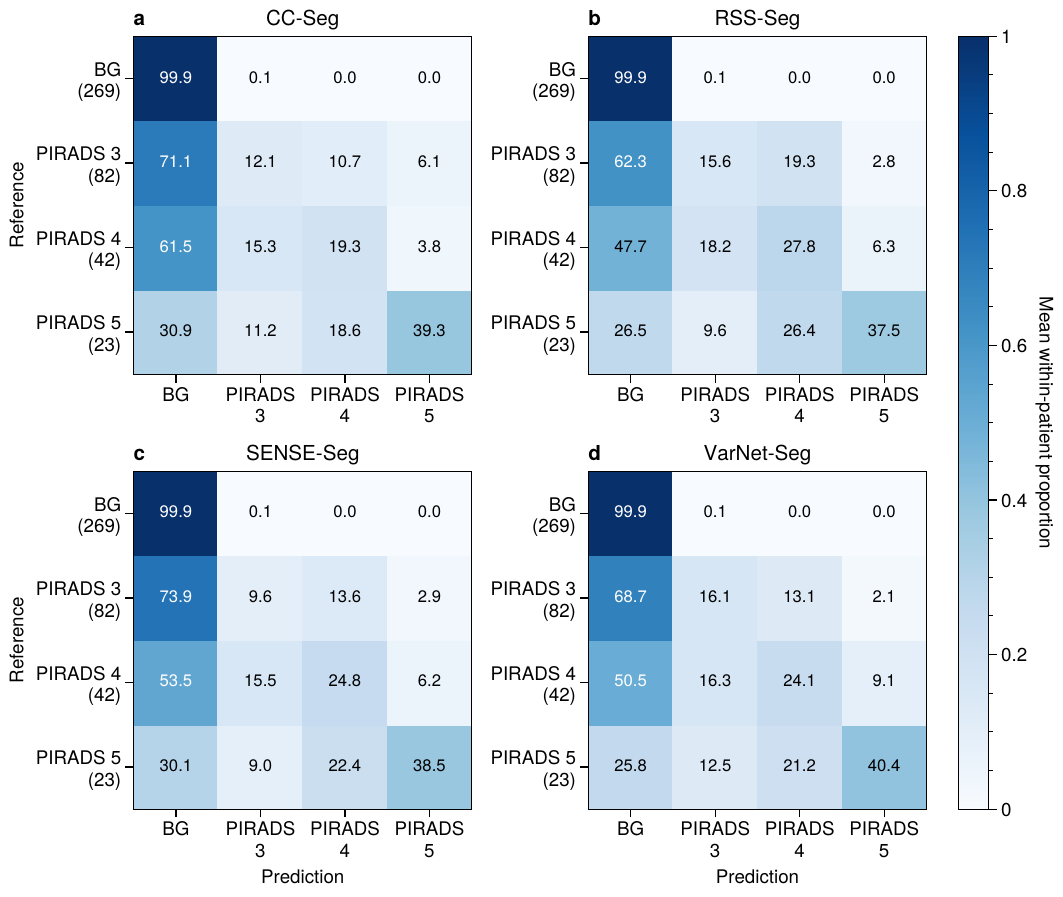}
    \caption{Confusion matrices with full sampling for CC-Seg, RSS-Seg, SENSE-Seg and VarNet-Seg. Values are percentages, normalized by reference class within each patient and averaged across patients with that class. Patient counts are given in parentheses. BG, background.}
    \label{fig:clean}
\end{figure}

Grade-specific Dice for PIRADS~4 and 5 is shown in Figure~S1 of the Supplement.
Across models, grade-specific Dice was higher for PIRADS~5 and lower for PIRADS~4 lesions. VarNet-Seg achieved 0.412 and 0.226, respectively, with full sampling.
The confusion matrices in Figure~\ref{fig:clean} showed the classification for the different PIRADS categories. Approximately half of the PIRADS~4 reference voxels were classified as background.

\subsection{Computational cost}
Table~\ref{tab:efficiency} summarizes the computational cost.
VarNet-Seg was slightly larger, with 0.41 million more parameters than the models with fixed reconstructions and a seven-fold longer inference time than SENSE-Seg. All four VarNet variants had the same parameter count and an inference time of approximately 29~ms per volume. For the frozen variants, training time includes both reconstruction pretraining and subsequent segmentation training.

\begin{table}[!htbp]
    \caption{Computational cost for training and inference measured across folds including parameter counts. Training time includes validation and, for the frozen variants, both reconstruction pretraining and segmentation training. Inference measures the GPU forward pass only.}
    \label{tab:efficiency}
    \setlength{\tabcolsep}{4pt}
    \begin{tabular}{@{}lrrr@{}}
        \toprule
        Model & \shortstack{Parameters\\(M)} & \shortstack{Training time\\(min)} & \shortstack{Inference time\\(ms/volume)} \\
        \midrule
        CC-Seg & 2.71 & 6.56 & 3.13 \\
        RSS-Seg & 2.71 & 30.12 & 3.85 \\
        SENSE-Seg & 2.71 & 27.23 & 4.15 \\
        VarNet-Seg & 3.12 & 60.05 & 29.18 \\
        VarNet-Seg (no DC) & 3.12 & 59.77 & 29.18 \\
        VarNet-Seg (frozen) & 3.12 & 58.20 & 29.17 \\
        VarNet-Seg (frozen, no DC) & 3.12 & 53.91 & 29.15 \\
        \bottomrule
    \end{tabular}
\end{table}

\FloatBarrier
\section{Discussion}
\label{sec:discussion}
In this study we trained a novel end-to-end model which includes a variational network for k-space processing and 3D image-space network for lesion segmentation and detection on multicoil prostate DWI. The proposed end-to-end model performed similarly to RSS-Seg at full sampling and achieved the highest PDice across undersampling factors, as shown in Figure~\ref{fig:masking}. Matched ablations supported contributions from both end-to-end training and data consistency.

Previous work showed that segmentation can be learned from undersampled MRI, such as in predicting cardiac structures directly from the data or by coupling reconstruction and segmentation networks~\cite{Schlemper2018,Sun2019}. Interestingly, the K2S challenge found weak or absent associations between reconstruction metrics and segmentation accuracy~\cite{Tolpadi2023}, showing that optimizing reconstruction fidelity is not the ideal method for improving automated analysis accuracy. 

Several prostate reconstruction studies have also investigated whether and how strong acceleration in general impacts the diagnostic value of prostate MRI. Johnson et al. evaluated image quality and lesion identification by radiologists after variational reconstruction of biparametric MRI~\cite{Johnson2022}, and a more recent multireader study of T2-weighted MRI found no statistically significant reduction in AUROC up to six-fold acceleration, although AUROC did decrease as acceleration increased~\cite{VanLohuizen2026}. However, such reader studies only demonstrate how a fixed reconstruction method performs under acceleration, not which reconstruction is most useful to an automated segmentation network. We address the latter by letting the segmentation objective shape the reconstruction during training.

The comparison with SENSE-Seg shows that access to complex multicoil input alone does not explain the gain under undersampling, as this control shares the virtual coils, sensitivity maps and segmentation U-Net with VarNet-Seg. Relative to that matched control, the improvement reflects the addition of learned refinement together with its auxiliary reconstruction objective. The four matched VarNet variants further distinguish reconstruction training strategy from data consistency. End-to-end training improved Dice under undersampling relative to reconstruction-pretrained, frozen frontends both with and without data consistency. Data consistency also improved Dice under both training strategies. These comparisons support contributions from adapting reconstruction to the segmentation objective and from constraining the learned corrections with the acquired coil data.

The benefits of the end-to-end model come at a cost in inference time, as reported in Table~\ref{tab:efficiency}. The Dice advantage over RSS-Seg was most pronounced with undersampled or noisy input.

Dice and patient-level AUROC characterize different aspects of the predictions. Dice measures spatial agreement in patients with reference lesions, whereas the patient-level score ranks examinations by predicted lesion volume, counting correct and misplaced predictions alike and ignoring their confidence. Patient-level rankings therefore depend on both lesion burden and false-positive predictions.
Van Lohuizen et al. found an analogous gap, where improved image similarity after T2-weighted reconstruction did not restore detection performance to that with full sampling~\cite{VanLohuizen2024}.

Our study has several limitations. First, undersampling and noise were simulated retrospectively on data that had already been phase-corrected, averaged and coil-compressed, with sensitivity maps and normalization from the fully sampled data. The experiments compare robustness under controlled conditions but do not reproduce all real-world acquisition effects.  Second, the cohort comes from one institution and one scanner model, with only 64 patients contributing to the primary Dice endpoint. Third, lesions were annotated by a single reader without histopathological confirmation.

The natural next step is therefore an independent multicenter study on prospectively undersampled DWI, evaluated with Dice, lesion-level detection and calibrated patient-level scores against histopathology or clinical follow-up. Only such a study, timed over the complete acquisition and processing workflow, can show whether the gain survives in a genuinely shorter examination.
\section{Conclusion}
Our study supports an end-to-end approach to multicoil DWI for more robust prostate lesion segmentation under undersampling. The matched ablations support contributions from both data-consistency and end-to-end training. Independent prospective studies should establish whether this robustness translates into clinical benefit with accelerated prostate DWI.

\backmatter
\section{Declarations}

\bmhead{Ethics approval and consent to participate}
The fastMRI Prostate source publication documents approval of data collection and sharing by the NYU Langone Health Institutional Review Board (S18-00412, S23-00146) and a waiver of informed consent~\cite{Tibrewala2024}.

\bmhead{Data and code availability}
The original MRI data and associated PIRADS ratings are available through the fastMRI Prostate release~\cite{Tibrewala2024}.

\bmhead{Competing interests}
The authors have no relevant conflicts of interest to disclose.

\bmhead{Funding}
Computing resources were provided by the German AI Service Center WestAI on the JURECA supercomputer at the J{\"u}lich Supercomputing Centre (JSC), Forschungszentrum J{\"u}lich~\cite{JURECA2021}.

\bmhead{Use of AI-assisted tools}
OpenAI Codex was used to assist with manuscript drafting, editing, and LaTeX preparation.

\bmhead{Supplementary information}
The Supplement provides additional details on the cohort, preprocessing, training and evaluation. Tables~\ref{tab:cohort}, \ref{tab:settings} and \ref{tab:conditions} summarize the cohort, training settings and sampling parameters. Figure~S1 shows Dice for PIRADS~4 and 5.
\FloatBarrier
\bibliography{references}
\clearpage
\section*{Supplementary information}
\numbered
\setcounter{section}{0}
\setcounter{subsection}{0}
\setcounter{figure}{0}
\setcounter{table}{0}
\setcounter{equation}{0}
\renewcommand{\thesection}{S\arabic{section}}
\renewcommand{\thesubsection}{\thesection.\arabic{subsection}}
\renewcommand{\thefigure}{S\arabic{figure}}
\renewcommand{\thetable}{S\arabic{table}}
\renewcommand{\theequation}{S\arabic{equation}}
\renewcommand{\theHsection}{supp.\arabic{section}}
\renewcommand{\theHfigure}{supp.\arabic{figure}}
\renewcommand{\theHtable}{supp.\arabic{table}}
\renewcommand{\theHequation}{supp.\arabic{equation}}

\section{Supplementary methods}

\subsection{Study cohort}

\begin{table}[!htbp]
    \caption{Study cohort. Examinations with lesions are classified by the highest annotated DWI PIRADS score.}
    \label{tab:cohort}
    \begin{tabular}{@{}p{0.68\linewidth}r@{}}
        \toprule
        Lesion category & Patients \\
        \midrule
        No annotated lesion & 130 \\
        PIRADS 3 & 75 \\
        PIRADS 4 & 41 \\
        PIRADS 5 & 23 \\
        \midrule
        Total & 269 \\
        \bottomrule
    \end{tabular}
\end{table}

\subsection{Data preparation}

GRAPPA reconstruction used a $5\times5$ kernel calibrated per slice from 32 central lines~\cite{Griswold2002}. Coil compression was calibrated once per examination using all slices~\cite{Zhang2013}. A low-resolution phase estimate from the central $32\times32$ k-space region, with a Hamming window, was subtracted from every coil data before complex arithmetic averaging across repetitions and diffusion directions~\cite{Eichner2015}. ESPIRiT maps used a calibration width of 32 lines and were phase-aligned to the low-resolution $b50$ SENSE image~\cite{Uecker2014}.

\subsection{Loss and training settings}

The Dice term was calculated as the average over foreground classes present in each training window~\cite{Milletari2016}. The voxel-mean categorical focal loss included background and used $\gamma=2$~\cite{Lin2017}. For VarNet-Seg and VarNet-Seg (no DC), the reconstruction loss was defined by the mean squared error between output magnitudes and the SENSE magnitudes of the augmented input with full sampling. Errors were normalized by the mean squared reference magnitude for each contrast and window within the segmentation crop, then averaged.
For VarNet-Seg (frozen) and VarNet-Seg (frozen, no DC), reconstruction pretraining used this magnitude NMSE alone with weight 1 and no segmentation loss. Reconstruction checkpoint selection used fixed equispaced masks at $R=2,4,6,8$ on the inner-validation patients, without added noise or spatial augmentation. Validation NMSE was averaged over windows within each patient, then over patients and accelerations, with equal weighting of both $b$-values and exclusion of padded slices.

After pretraining, all VarNet parameters, including residual scales and any active data-consistency weights, were frozen. A fresh segmentation U-Net was trained with the original segmentation sampling, augmentation and loss, without an auxiliary reconstruction loss. Each stage used the same optimizer and learning-rate schedule, a maximum of 100 epochs and early-stopping patience 30. Data consistency was enabled throughout both stages for VarNet-Seg (frozen) and disabled throughout both stages for VarNet-Seg (frozen, no DC). Outer-test patients were excluded from both stages' fitting and checkpoint selection.

Table~\ref{tab:settings} gives the loss weights and training settings.

\begin{table}[!htbp]
    \centering
    \small
    \renewcommand{\arraystretch}{1.12}
    \caption{Model and training settings.}
    \label{tab:settings}
    \begin{tabular}{@{}p{0.24\linewidth}p{0.71\linewidth}@{}}
        \toprule
        Component & Setting \\
        \midrule
        Image dimensions & $100\times100$-pixel reconstruction matrix, central $64\times64$-pixel crop for segmentation and 24-slice input windows. \\
        VarNet module & Six cascades with separate weights, 24 channels and two residual blocks with in-plane dilations of 1 and 2, without pooling. Final layers initialized to zero, residual scales to 0.1 and data-consistency weights to 0.5. \\
        Segmentation U-Net & Channels 24, 48, 96 and 192, with strides $(1,1,1)$, $(1,2,2)$ and $(2,2,2)$. Two residual units per level, group normalization with two groups and LeakyReLU with slope 0.1. A linear layer maps 24 output features to four classes. \\
        Batch and sampling & Batch size 16. Windows containing target lesions receive half the total sampling probability. \\
        Loss weights & Dice loss 0.7 and focal loss 0.15. Reconstruction loss 0.1 for VarNet-Seg and VarNet-Seg (no DC). \\
        Optimization & AdamW, learning rate $2\times10^{-4}$, weight decay 0.01 and gradient clipping at 1.0. Cosine decay to $10^{-6}$ over 100 epochs, with early-stopping patience 30. \\
        Spatial augmentation & Flips, $90^\circ$ rotations, rotations up to $\pm15^\circ$, translations up to $\pm5$ pixels and zoom 0.85--1.25. Each applied with probability 0.5 per batch. \\
        Signal augmentation & Gamma 0.7--1.5, undersampling at factors 2, 4, 6 or 8, and noise at SNRs of 20, 10 or 5~dB, each applied with probability 0.5. Random and regularly spaced undersampling were used equally often. \\
        Implementation & PyTorch 2.11, MONAI 1.5.1 and NVIDIA H100 GPUs. FP32 reconstruction and probability averaging, with a BF16 U-Net. \\
        \bottomrule
    \end{tabular}
\end{table}

\section{Evaluation and statistical analysis}

\subsection{Undersampling and noise}

Table~\ref{tab:conditions} lists the sampling parameters. The central band was retained in addition to regularly spaced phase-encoding lines. Sampling patterns were shared across coils, contrasts and slices within an examination and across models. Noise-only experiments used SNRs from 30 to $-20$~dB in 5~dB steps. Combined experiments used undersampling factors 2, 4, 6, 8, 10, 12 and 16 at SNRs of 20, 10 and 5~dB.

\begin{table}[!htbp]
    \centering
    \small
    \renewcommand{\arraystretch}{1.12}
    \caption{Sampling parameters for the 100 phase-encoding lines. Retained line counts include the central band.}
    \label{tab:conditions}
    \begin{tabular}{@{}rrr@{}}
        \toprule
        Undersampling factor & Central lines & Retained lines \\
        \midrule
        1 & --- & 100 \\
        2 & 4 & 52 \\
        4 & 8 & 31 \\
        6 & 5 & 20--21 \\
        8 & 4 & 16 \\
        10 & 3 & 12--13 \\
        12 & 2 & 10--11 \\
        16 & 2 & 8--9 \\
        \bottomrule
    \end{tabular}
\end{table}

Complex Gaussian noise was added before undersampling, with variance $\sigma^2=C\bar P/10^{\mathrm{SNR}/10}$.
Here, $C$ is the number of coils per contrast and $\bar P$ is the mean squared value of the real and imaginary k-space samples with full sampling, pooled across coils, slices and both contrasts.

\subsection{Statistical comparisons}

\begin{figure}[!htbp]
    \centering
    \includegraphics[width=\linewidth]{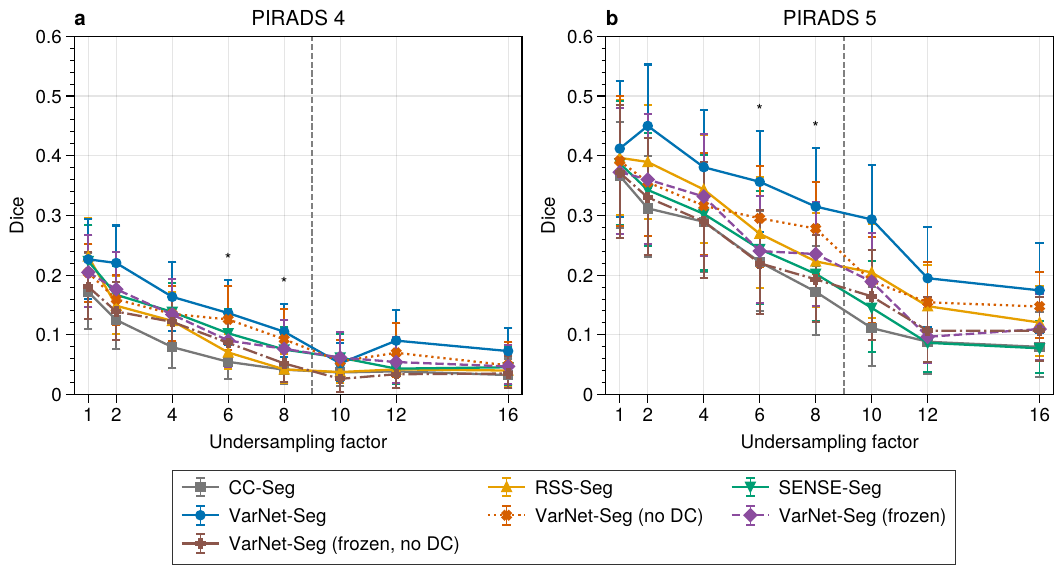}
    \caption{Grade-specific Dice with undersampling for all seven models, including VarNet-Seg (frozen) and VarNet-Seg (frozen, no DC). (a) PIRADS~4 and (b) PIRADS~5. Error bars show 95\% bootstrap confidence intervals. Asterisks indicate VarNet-Seg--RSS-Seg comparisons ($^{*}p<0.05$, $^{**}p<0.01$, $^{***}p<0.001$).}
    \label{fig:supp-grade}
\end{figure}

Confidence intervals were calculated from the 2.5th and 97.5th percentiles of 10,000 patient bootstrap samples stratified by fold and highest reference grade~\cite{Efron1994}.
Two-sided tests compared VarNet-Seg--RSS-Seg differences with their null-centered bootstrap distribution, using an add-one correction.
Comparisons were grouped by the panels in Figures~2a and 4b--d of the main manuscript and Figure~\ref{fig:supp-grade}.

\FloatBarrier
\section{Grade-specific Dice}

\FloatBarrier
\end{document}